\documentclass[sigconf,nonacm]{acmart}
\AtBeginDocument{%
  }
\usepackage{multirow,multicol}
\usepackage{fancyhdr}
\usepackage{subcaption} 
\usepackage{booktabs}
\usepackage{multirow}
\usepackage[table]{xcolor}
\usepackage{colortbl}
\usepackage{pifont}
\usepackage{makecell}
\graphicspath{{./}{figs/}{../}}
\newcommand{\figcol}{\linewidth}      
\newcommand{\figspan}{0.85\linewidth} 
\newcommand{\figplot}{0.8\linewidth}  
\newcommand{\tabsota}{0.7\textwidth}  
\newcommand{\tabhalf}{\linewidth}     
\renewcommand\footnotetextcopyrightpermission[1]{}

\definecolor{cleanrow}{RGB}{245, 245, 245}
\definecolor{mixedrow}{RGB}{255, 235, 235}
\definecolor{tttrow}{RGB}{230, 244, 255}
\definecolor{avgrow}{RGB}{245, 240, 255}
\definecolor{bestcolor}{RGB}{220, 20, 60}

\acmConference[Conference acronym 'MM]{}{}{}

\acmSubmissionID{1931}

\begin{document}

\title{Multi-Modal Hyper-Graph Fusion for Low-Light Crowd Counting}


\author{Hao-Yuan Ma}
\affiliation{%
  \institution{School of Computer Science and Technology, Soochow University}
  \city{Suzhou}
  \country{China}}

\author{Li Zhang}
\affiliation{%
  \institution{School of Computer Science and Technology, Soochow University}
  \city{Suzhou}
  \country{China}}

\author{Yushi Qiu}
\affiliation{%
  \institution{School of Computer Science and Technology, Soochow University}
  \city{Suzhou}
  \country{China}}

\author{Jie Gao}
\affiliation{%
  \institution{School of Computer Science and Technology, Soochow University}
  \city{Suzhou}
  \country{China}}

\author{Yan Zhang}
\affiliation{%
  \institution{School of Computer Science and Technology, Soochow University}
  \city{Suzhou}
  \country{China}}

\author{Bangjun Wang}
\affiliation{%
  \institution{School of Computer Science and Technology, Soochow University}
  \city{Suzhou}
  \country{China}}







\renewcommand{\shortauthors}{Ma et al.}

\begin{abstract}
    Crowd counting is a fundamental task in computer vision. However, crowd counting in low-light environments remains largely underexplored, despite its practical importance in the real world. 
    Existing methods mainly focus on well-lit scenes or rely on single-modality Red-Green-Blue (RGB) representations, which often become unreliable under extreme darkness and complex non-uniform illumination. To handle this problem, we construct three new low-light crowd counting benchmarks, which consist of two synthetic datasets, SHA\_Dark and SHB\_Dark, and a real-world benchmark LC-Crowd (Low-light Crowd Dataset). 
    Inspired by Retinex-based physical modeling, we introduce depth and Canny edge cues as complementary geometric and structural priors to enhance the intrinsic reflectance representation under low-light conditions. 
    We propose a Multi-Modal Hyper-Graph Fusion module, which formulates RGB appearance, depth geometry, and edge structure cues as nodes in a unified hyper-graph and explicitly captures their high-order complementary relationships via dynamic hyperedge construction and message passing. Furthermore, to adaptively allocate computation in dense prediction, we propose a Deformable Rectangular Sparse Attention (DRSA) module, which concentrates computation on informative regions through anchor-aware estimation and adaptive rectangular window modeling.
 Based on these designs, we develop a unified Low-Light Counting Network (LCNet) for robust low-light crowd counting. Extensive experiments on three benchmarks demonstrate that the proposed method achieves the best overall performance against existing state-of-the-art (SOTA) methods. The code is in the supplementary material. The datasets will be made public upon acceptance.
\end{abstract}





\maketitle

\section{Introduction}
Crowd counting is a fundamental task in computer vision, with broad applications in intelligent surveillance \cite{surveillance}, public safety \cite{PublicSafety}, urban management \cite{UrbanManagement}, and large-scale event monitoring \cite{crowdMonitoring}. In recent years, data-driven crowd counting methods have achieved remarkable progress in well-lit environments \cite{vmambacc,M2PLNet,FGENet,P2PNet}. However, many real-world dense crowd scenarios, such as concerts, night markets, evening gatherings, and festival events, often occur under low-light conditions, where robust crowd perception is even more critical for risk warning and emergency response. Unfortunately, insufficient illumination, severe sensor noise, low contrast, and the loss of structural details cause substantial visual degradation, under which standard RGB counting models trained on daytime data tend to fail. Therefore, low-light crowd counting is an important problem that, despite its practical significance, remains insufficiently explored.

Existing crowd counting methods can be broadly categorized into density-based and point-based paradigms. Density-based approaches estimate a continuous density distribution and obtain the crowd number via integration \cite{MCNN}, while point-based methods \cite{P2PNet,CLTR} directly predict individual head locations or point sets, usually achieving better localization in dense scenes. Although both paradigms perform strongly under normal illumination, they remain sensitive to low light. To cope with darkness, most existing solutions either learn from RGB appearance alone or adopt a sequential \emph{enhance-then-count} pipeline, in which a low-light enhancement model is first applied to improve visibility and the enhanced image is then fed into a standard counter. However, such designs treat enhancement and counting as two independent tasks, and the enhancement stage may amplify sensor noise, over-smooth crowd textures, or remove the very structural details that counting relies on, thereby accumulating errors across stages. This suggests that, rather than restoring a visually pleasing image at the \emph{pixel} level, what truly matters is recovering reliable, illumination-robust crowd representations at the \emph{feature} level.

To formalize this intuition, we resort to Retinex theory \cite{retinex}, which models an observation as $I=R\odot L$, where the intrinsic reflectance $R$ is corrupted by an unfavorable illumination $L$ under low light. We extend this decomposition from the pixel space to the feature space and cast low-light crowd counting as a problem of \emph{reflectance re-calibration}: recovering the illumination-robust, reflectance-related component of the degraded RGB feature by leveraging complementary priors. The central questions then become \emph{which} priors to use for calibration and \emph{how} to perform the calibration reliably.

We answer the first question by introducing depth and Canny edge cues as calibration anchors. The key observation is that low-light corruption is \emph{cue-dependent}: while RGB appearance is highly sensitive to illumination, the underlying scene geometry and object boundaries are comparatively illumination-invariant. Depth encodes 3D layout that stays stable in dark regions, and edge responses preserve head and body contours even when texture is washed out by underexposure. These two priors therefore provide reliable, illumination-robust anchors for re-calibrating the degraded RGB reflectance. We note that throughout this paper the term \emph{multi-modal} refers to the joint use of RGB appearance together with geometry (depth) and structure (edge) cues, where depth and Canny are illumination-robust visual priors derived from or aligned with the RGB observation rather than signals from extra physical sensors. Nevertheless, under extreme darkness these priors are not uniformly reliable either.

This brings us to the second and more critical question. In severely dark scenes, any single cue can fail locally: depth degenerates into noise in textureless dark areas, and edges break up under heavy noise. If calibration is performed through \emph{pairwise} relations, as in standard attention or graph convolution, a single corrupted neighbor can readily contaminate a node. To be robust, a node should instead aggregate evidence from \emph{a group} of mutually consistent neighbors simultaneously, so that the calibration tolerates the failure of any individual cue. This is precisely what a hyperedge offers, namely a single edge connecting an arbitrary subset of nodes. Motivated by this, we propose a Multi-Modal Hyper-Graph Fusion module that connects the appearance-, geometry-, and structure-aware tokens within a unified hyper-graph and performs high-order, group-wise reflectance re-calibration through dynamic hyperedge reasoning, going beyond the pairwise relations of conventional fusion. After such high-order calibration, the features become more reliable, yet the foreground that actually carries crowd information remains highly sparse in nighttime scenes.

This sparsity motivates our last design. Applying dense attention over the whole feature map is not only computationally redundant but also tends to draw in background noise. We therefore introduce a Deformable Rectangular Sparse Attention (DRSA) mechanism that estimates the foreground anchor number per window through granularity filtering and performs deformable rectangular attention only on the selected anchors, thereby adaptively allocating computation to informative foreground regions. We position DRSA as a \emph{computation-adaptive} design that places computation where it matters, rather than one that simply maximizes frame rate. Built upon these designs, we develop a unified Low-Light Counting Network (LCNet) for robust low-light crowd counting. In addition, since dedicated low-light benchmarks are still scarce, we establish three new benchmarks, consisting of two synthetic datasets, SHA\_Dark and SHB\_Dark, and a real-world benchmark LC-Crowd (Low-light Crowd Dataset), where the synthetic ones provide controllable degradation for standardized evaluation and the real-world one captures diverse nighttime scenes with large scale variation.

The main contributions of this work are summarized as follows:

\begin{itemize}
    \item To address the scarcity of dedicated benchmarks, we establish three low-light crowd counting benchmarks, including two synthetic datasets, SHA\_Dark and SHB\_Dark, and one real-world benchmark LC-Crowd.
    \item To address the unreliability of any single cue under extreme darkness, we cast low-light crowd counting as reflectance re-calibration and propose a Multi-Modal Hyper-Graph Fusion module that performs high-order, group-wise calibration over appearance, geometry, and structure cues, going beyond pairwise fusion.
    \item To address the spatial sparsity of foreground in nighttime scenes, we propose a Deformable Rectangular Sparse Attention module that adaptively allocates computation to informative regions for dense prediction.
    \item We develop a unified network, LCNet, and extensive experiments show that it achieves the best overall performance on all three benchmarks and remains competitive against recent multi-modal counting methods.
\end{itemize}

\section{Related Work}

\subsection{Crowd Counting Methods}
Existing crowd counting methods can be broadly divided into density-based and point-based paradigms. Density-based methods estimate crowd numbers by regressing continuous density maps, with representative works including MCNN \cite{MCNN}, CSRNet \cite{CSRNet}, SANet \cite{SASNet}, GauNet \cite{GauNet}, CAN \cite{CANNet}, and Crowddiff \cite{crowddiff}. Point-based methods directly predict head centers or point sets through proposal and matching mechanisms, and typical methods include P2PNet \cite{P2PNet}, FGENet \cite{FGENet}, PET \cite{PET}, M2PLNet \cite{M2PLNet}, APGCC \cite{APGCC}, and VMambaCC \cite{vmambacc}. While both paradigms have achieved promising performance, they are mainly developed under normal illumination and often degrade in low-light scenes due to underexposure, weak contrast, sensor noise, and blurred boundaries. 

Existing solutions for low-light crowd counting mainly follow two directions. One direction relies on dedicated sensing hardwaresuch as thermal, infrared, RGB-T \cite{RGB_CC,RGBT_TMM} systems, to compensate for the weakness of RGB imagery under challenging illumination. These methods can provide illumination-insensitive cues and improve perception robustness but usually require specialized acquisition devices and are less applicable to common surveillance settings. However, these methods usually depend on specialized devices and are less suitable for common RGB surveillance scenarios. The other direction adopts an enhance-then-count pipeline, where low-light enhancement methods, such as LIME \cite{LIME}, LLNet \cite{LLNet}, Zero-DCE \cite{Zero-DCE}, and Retinex-based methods \cite{Retinex1,Retinex2,retinexformer}, are first applied to improve visibility before counting. However, decoupling enhancement and counting may amplify noise, distort textures, or remove details important for accurate counting.

Some recent studies have also explored crowd analysis in adverse conditions such as haze and rain \cite{awcc}, showing that degradation significantly alters crowd features and hinders direct transfer from clear-scene models. Nevertheless, low-light crowd counting remains underexplored, and dedicated benchmarks and tailored methods are still scarce.

\subsection{Crowd Counting Datasets}
Benchmark datasets have played an important role in the development of crowd counting. Early datasets mainly focus on RGB images under normal illumination. Mall \cite{Mall} is one of the earliest benchmarks, containing surveillance frames captured from a fixed shopping mall camera. With limited scene variation and moderate density changes, it is widely used for evaluating early counting models.
ShanghaiTech \cite{MCNN} is a standard benchmark in crowd counting. It consists of two parts: Part A with dense Internet images showing severe occlusion and large scale variation, and Part B with relatively sparse urban street scenes. Its diversity in crowd density and scene complexity makes it widely used for evaluating counting accuracy and generalization.
JHU-Crowd++ \cite{JHU} and NWPU-Crowd \cite{NWPU} further increase the scale and difficulty of crowd counting benchmarks. JHU-Crowd++ contains unconstrained scenes with variations in viewpoint, weather, illumination, and background, while NWPU-Crowd provides large-scale images ranging from sparse to extremely dense crowds with severe occlusion and perspective distortion. 
However, the above datasets are mostly collected under normal daytime illumination and rely on RGB imagery. For low-light scenarios, RGBT-CC \cite{IADM} introduces aligned RGB and thermal images, providing complementary cues for crowd counting in nighttime or poor-visibility conditions. Nevertheless, it depends on specialized sensors and is different from common RGB surveillance settings.

Overall, existing datasets have evolved from early fixed-camera benchmarks to large-scale unconstrained scenes, but dedicated benchmarks for low-light crowd counting remain scarce.
\begin{figure}[h]
    \centering
    \includegraphics[width=\figcol]{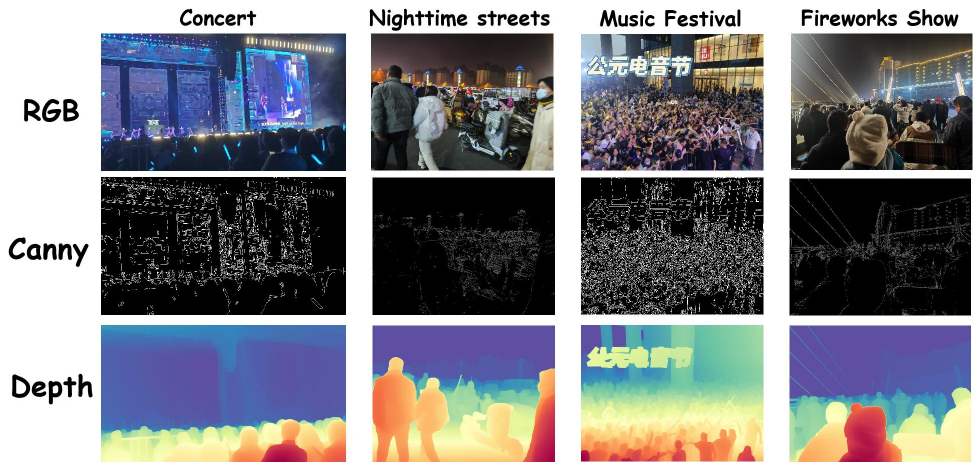}
    \caption{Visual examples from the Low-Light Crowd Counting Dataset. The columns illustrate diverse real-world low-light scenarios. From top to bottom, the rows display the original RGB images, the corresponding canny edge maps, and the depth maps, highlighting the multi-modal representations provided by the dataset.}
    \label{fig:llc_dataset}
\end{figure}

\section{Benchmark}
\subsection{Overview}
To facilitate low-light crowd counting, we establish three dedicated multi-modal benchmarks: two synthetic datasets, SHA\_Dark and SHB\_Dark, and one real-world dataset, LC-Crowd. 
In contrast to conventional crowd counting datasets, which mainly emphasize well-lit scenes, our benchmark targets severely underexposed settings characterized by uneven illumination. Each sample includes four components: an RGB image, a depth prior, a canny edge prior, and crowd counting annotations. This design explicitly supports cross-modal reasoning over appearance, geometry, and structure. Compared with conventional single-modality benchmarks \cite{MCNN,JHU,NWPU}, the proposed protocol provides complementary modal cues for robust perception under poor illumination and enables the study of multi-modal fusion strategies for low-light crowd analysis.

\begin{figure}
    \centering
    \includegraphics[width=\figplot]{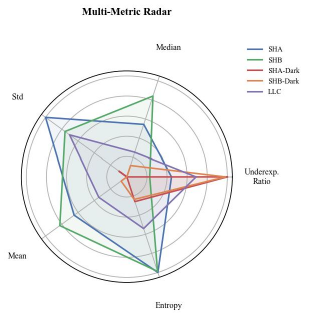}
    \caption{Radar chart comparison of five illumination-related statistics (mean brightness, median brightness, standard deviation, underexposure ratio, and entropy) across the original well-lit benchmarks (SHA, SHB) and the proposed low-light benchmarks (SHA\_Dark, SHB\_Dark, and LC-Crowd). ``LC-Crowd'' denotes our real-world Low-light Crowd dataset. The clear gap between the well-lit and low-light datasets confirms the severity and distinctiveness of the proposed benchmarks.}
    \label{fig:dataset_plot_radar}
\end{figure}

\subsection{LC-Crowd Dataset}
Existing crowd counting datasets lack sufficient coverage of diverse real-world low-light scenarios. 
To bridge this gap, we construct the LC-Crowd dataset, a new benchmark specifically designed for crowd counting in dark environments. As shown in Fig.~\ref{fig:llc_dataset}, the dataset contains 529 images collected from diverse nighttime scenes, including concerts, fireworks show, and evening gatherings, covering a broad range of crowd scales and illumination patterns. 
A notable characteristic of this dataset is its large scale variation, with image resolutions ranging from 600 $\times$ 400 to 6499 $\times$ 4763 pixels. In total, the dataset includes 30,932 manually annotated head points, with crowd counts varying from 1 to 1,984, and an average of 58.47 persons per image. 

As illustrated in Fig.~\ref{fig:dataset_plot_radar}, the dataset presents substantial real-world illumination complexity, with a mean brightness of 69.40, a standard deviation of 29.80, and brightness values ranging from 5.53 to 183.62. This wide dynamic range reflects severe non-uniform illumination and creates significant challenges for conventional crowd counting methods.

For annotation, we manually label the center location of each individual head as the counting ground truth. 
Notably, to annotate this low-light dataset, we developed a dedicated low-light annotation system. The system first adaptively enhances the images using Zero-DCE \cite{Zero-DCE}, which annotators can further improve visibility under low-light conditions by adjusting contrast, illumination intensity, and other parameters. 
In addition, it can invoke FGENet \cite{FGENet} which is pretrained on the JHU-Crowd++ dataset \cite{JHU}, to provide auxiliary annotations for the images. 
Beyond the RGB image and point annotations, we further associate each sample with two auxiliary modalities: a depth modality, which provides illumination-insensitive geometric cues, and a canny modality, which captures boundary and structural information that remains informative even when texture details are weakened by darkness. 
Concretely, the depth maps are generated by the monocular depth estimator Depth Anything V2 \cite{depthanythingv2} using its publicly released pretrained checkpoint, applied directly to the RGB images without any task-specific fine-tuning; the predicted inverse-depth maps are then min-max normalized to $[0,1]$. The Canny maps are extracted with the standard Canny detector on the grayscale image, where the low and high hysteresis thresholds are set to 50 and 150, respectively, after a $3\times3$ Gaussian smoothing. We keep this generation pipeline identical across all three benchmarks so that the auxiliary cues are reproducible and the comparison remains consistent.

\begin{figure*}[t]
    \centering
    \includegraphics[width=\figspan]{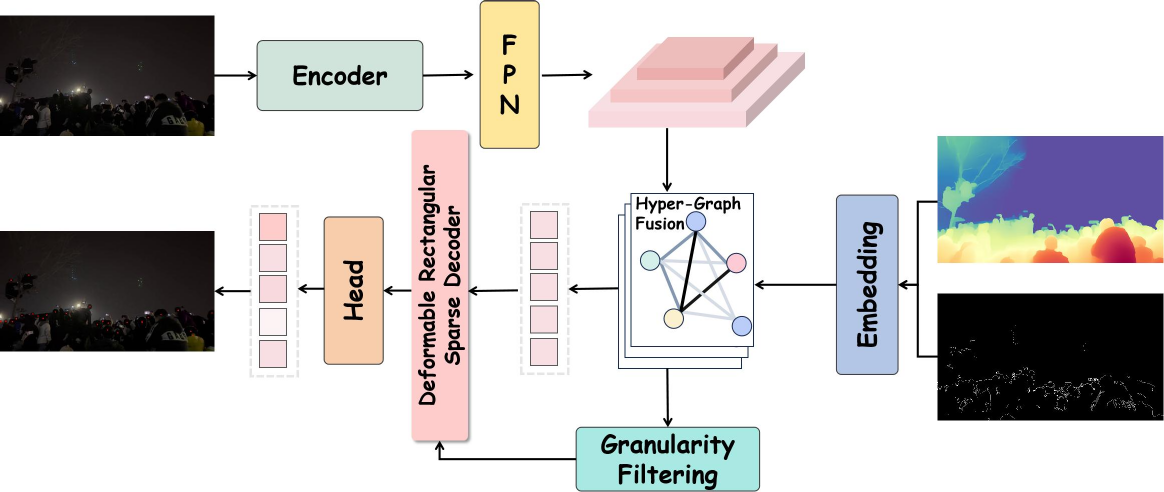}
    \caption{The overall architecture of the proposed LCNet for low-light crowd counting. The low-light RGB image is first encoded by the encoder and FPN to obtain hierarchical features. In parallel, depth and canny modalities are projected into a shared embedding space and integrated with RGB features via Hyper-Graph Fusion. The fused features are then refined by Granularity Filtering and processed by the Deformable Rectangular Sparse Decoder for efficient dense representation learning. The final prediction head produces the counting output.}
    \label{fig:overview}
\end{figure*}

\subsection{Synthetic Low-light Benchmarks}
To complement real-world evaluation, we further construct two synthetic low-light datasets, SHA\_Dark and SHB\_Dark, based on ShanghaiTech Part A and Part B \cite{MCNN}. By applying gamma correction and exposure reduction to the original well-lit images, these datasets simulate realistic low-light conditions while retaining normal-light references for controlled comparison. As shown in Fig.~\ref{fig:dataset_plot_radar}, both datasets differ clearly from normal-light benchmarks in brightness, underexposure ratio, and entropy, demonstrating their effectiveness in modeling challenging low-light scenarios. In addition, to match the proposed multi-modal setting, we extend both datasets with corresponding depth and canny modalities, enabling a more systematic analysis of modality complementarity under illumination degradation. Detailed brightness and count statistics of all three benchmarks are summarized in Tables~\ref{tab:brightness_comparison} and \ref{tab:dataset_statistics}.

\begin{table}[t]
\centering
\caption{Brightness statistics of the original well-lit datasets and the proposed low-light benchmarks.}
\label{tab:brightness_comparison}
\resizebox{\linewidth}{!}{
\begin{tabular}{lccccc}
\hline
Dataset & Mean & Median & Std & Underexp. Ratio & Entropy \\
\hline
SHA      & 96.72  & 84.15  & 64.71 & 39.9\% & 7.58 \\
SHB      & 112.58 & 111.85 & 54.44 & 24.5\% & 7.56 \\ \hline
SHA\_Dark & 38.30  & 33.27  & 25.89 & 79.4\% & 6.29 \\
SHB\_Dark & 44.63  & 44.36  & 21.79 & 78.0\% & 6.25 \\
LC-Crowd  & 69.40  & 57.02  & 52.03 & 57.3\% & 6.78 \\
\hline
\end{tabular}
}
\end{table}

\begin{table}[t]
\centering
\caption{Statistics of the proposed low-light crowd counting benchmarks.}
\label{tab:dataset_statistics}
\resizebox{\linewidth}{!}{
\begin{tabular}{l c c c c c c c}
\toprule
\multirow{2}{*}{Dataset} & \multirow{2}{*}{\#Images} & \multirow{2}{*}{Train/Test} & \multirow{2}{*}{Avg. Resolution} & \multicolumn{4}{c}{Count Statistics} \\
\cmidrule(lr){5-8}
 &  &  &  & Total & Min & Avg & Max \\
\midrule
SHA\_Dark & 482 & 300/182 & 589 $\times$ 868 & 241,677 & 33 & 501.4 & 3,139 \\
SHB\_Dark & 716 & 400/316 & 768 $\times$ 1024 & 88,488 & 9 & 123.6 & 578 \\
LC-Crowd & 529 & 449/80 & 600$\times$400 $\sim$ 6499$\times$4763 & 30,932 & 1 & 58.47 & 1,984 \\
\bottomrule
\end{tabular}
}
\end{table}
As reported in Table~\ref{tab:brightness_comparison}, the synthetic transformation substantially darkens the source data: the mean brightness drops from $96.72$/$112.58$ (SHA/SHB) to $38.30$/$44.63$ (SHA\_Dark/SHB\_Dark), the underexposure ratio rises above $78\%$, and the entropy decreases, confirming a significant and consistent low-light degradation. In contrast, LC-Crowd shows a higher mean brightness ($69.40$) yet a still-high underexposure ratio ($57.3\%$), indicating that it captures non-uniform real nighttime illumination rather than uniformly darkened scenes. Table~\ref{tab:dataset_statistics} further shows that the three benchmarks are complementary in scale and density: SHA\_Dark is highly congested (avg. $501.4$), SHB\_Dark is sparser street-view (avg. $123.6$), and LC-Crowd spans an extremely wide resolution range with diverse real-world densities.

\section{Method}

\subsection{Framework Overview}
We propose LCNet, a unified multi-modal framework for low-light crowd counting, as shown in Fig.~\ref{fig:overview}. Given a low-light RGB image together with its corresponding depth and canny modalities, LCNet aims to learn robust crowd representations under severe illumination degradation.

Inspired by Retinex theory \cite{retinex}, we extend the decomposition idea from the pixel space to the feature space. We assume that the feature representation can be disentangled into reflectance-related and illumination-related components:
\begin{align}
I=R\odot L\\
F = F^{r} \odot F^{l}    
\end{align}

where $I$ is the input image, $R$ and $L$ denote the reflectance and illumination. For the feature map $F$, we can decompose into reflectance $F^{r}$ and illumination components $F^{l}$.

Our goal is to improve intrinsic reflectance-related features $F^{r}$ by introducing complementary signals from additional modalities. Under low-light conditions, the illumination-related component in the RGB modality is often degraded, which in turn weakens the overall feature representation. To alleviate this issue, we exploit auxiliary-modal features $F^{m}$ to compensate for the missing structural and detail information, yielding a more robust representation:
\begin{equation}
F \approx (F^{r}+ F^{m}) \odot \bar{F}^{l},
\end{equation}
where $\bar{F}^{l}$ denotes the component related to degraded illumination under low-light conditions, and $F^{m}$ represents complementary information provided by the auxiliary modalities.

Specifically, the RGB image is first encoded by a backbone network and a feature pyramid to obtain hierarchical multi-scale features, while the depth and Canny cues are embedded as complementary geometric and structural priors.
To effectively exploit cross-modal complementarity, we introduce a Multi-Modal Hyper-Graph Fusion module. It first aggregates the depth and edge priors into an auxiliary reflectance prototype, fuses it with the RGB feature, and then performs high-order relational reasoning by constructing a token-level hyper-graph within local windows, in which the appearance-, geometry-, and structure-aware tokens are connected through dynamic hyperedges.
On top of the fused representation, we further design a Deformable Rectangular Sparse Attention (DRSA) module to adaptively allocate computation in dense prediction via anchor-aware estimation and deformable rectangular context modeling. 
Finally, the refined features are fed into a point-based counting head for crowd localization and counting.
Throughout this paper, we use the term \emph{multi-modal} to denote the joint use of RGB appearance together with the depth geometry and edge structure cues; depth and Canny are treated as illumination-robust visual priors derived from (or aligned with) the RGB observation rather than as signals from additional physical sensors.

\subsection{Multi-Modal Hyper-Graph Fusion}

In the low-light environment, relying solely on RGB appearance features becomes unreliable for low-light crowd counting.
To alleviate this issue, we introduce \emph{depth} and \emph{canny edge} priors as complementary geometric and structural cues to assist in recovering more robust reflectance-aware representations.
Specifically, we propose a Multi-Modal Hyper-Graph Fusion module, which explicitly models the high-order interactions among RGB appearance, depth geometry, and edge structure.

\noindent\textbf{Why a hyper-graph?}
A standard pairwise graph or self-attention models relations as edges connecting exactly two nodes, so each node is calibrated by a weighted sum of \emph{individual} neighbors. Under severe low light, however, any single cue can be locally corrupted (e.g., noisy depth in textureless dark areas or fragmented edges under heavy noise), which makes pairwise calibration fragile. A hyper-graph generalizes an edge into a \emph{hyperedge} that connects an arbitrary subset of nodes, allowing a center node to aggregate evidence from a \emph{group} of mutually consistent neighbors at once. This group-wise, high-order aggregation is more robust to the failure of any individual neighbor, which is exactly the regime we face in low-light scenes. We formally define the hyper-graph and empirically verify this advantage against pairwise alternatives in Sec.~\ref{sec:ablation} (Table~\ref{tab:fusion_strategy}).

\noindent\textbf{Hyper-graph definition.}
We denote the hyper-graph as $\mathcal{G}=(\mathcal{V},\mathcal{E},\mathbf{H})$, where $\mathcal{V}$ is the set of $N$ nodes (each node is a spatial token within a local window of the multi-modal fused feature), $\mathcal{E}$ is the set of $E$ hyperedges, and $\mathbf{H}\in\mathbb{R}^{N\times E}$ is the incidence matrix that records which nodes belong to which hyperedge. Each hyperedge groups a center token with its most consistent neighbors, so that message passing over $\mathbf{H}$ realizes the high-order calibration described above.

Given the encoded RGB feature map
\begin{equation}
F^{rgb} \in \mathbb{R}^{B \times C \times H \times W},
\end{equation}
we first align the auxiliary modalities, namely the Canny map $M^{can}$ and the depth map $M^{dep}$, to the same spatial resolution as the RGB feature. Depending on the modality setting, the auxiliary input is defined as:
\begin{equation}
G^{can} = \sigma(\psi_{can}(M^{can})), \qquad
G^{dep} = \sigma(\psi_{dep}(M^{dep})),
\end{equation}
\begin{equation}
\tilde{M}^{can} = G^{can} \odot M^{can}, \qquad
\tilde{M}^{dep} = G^{dep} \odot M^{dep},
\end{equation}
\begin{equation}
F^{m} = \phi([\tilde{M}^{can}; \tilde{M}^{dep}]), \quad F^{m} \in \mathbb{R}^{B \times C \times H \times W},
\end{equation}
where $[\cdot;\cdot]$ denotes channel-wise concatenation, $\odot$ denotes element-wise multiplication, and $\sigma(\cdot)$ is the Sigmoid activation function. 
$\psi_{can}(\cdot)$ and $\psi_{dep}(\cdot)$ denote the gating networks for the canny edge modality and the depth modality are used to generate adaptive gating maps $G^{can}$ and $G^{dep}$. 
The modulated features $\tilde{M}^{can}$ and $\tilde{M}^{dep}$ are then concatenated and fed into $\phi(\cdot)$, which is implemented by two convolution layers followed by nonlinear activations.

The feature map $F^{m}$ serves as a reflectance prior which highlights informative foreground patterns while suppressing illumination-sensitive noise. To obtain an explicit foreground indicator that guides hyperedge construction, we compute a scalar foreground confidence at each spatial position by averaging the prior feature over channels:
\begin{equation}
 w_i = \frac{1}{C}\sum_{c=1}^{C} F^{m}_{c,i},\quad w_i\in W, 
\end{equation}
where $C$ is the number of channels, $F^{m}_{c,i}$ is the response of the $c$-th channel at the $i$-th spatial location, and $w_i$ indicates the foreground confidence of the $i$-th location. A larger $w_i$ implies that the corresponding token is more likely to belong to an informative crowd region.

After obtaining the reflectance prior feature, we concatenate $F^{m}$ with the RGB feature and project the concatenated representation into a unified embedding space:
\begin{equation}
F^{fuse} = \Psi([F^{rgb}; F^{m}]),
\end{equation}
where $\Psi(\cdot)$ is a $1 \times 1$ convolution. In this way, the original RGB appearance is enhanced by modality-derived structural and geometric cues prior to relational reasoning.

\begin{figure}[h]
    \centering
    \includegraphics[width=\figcol]{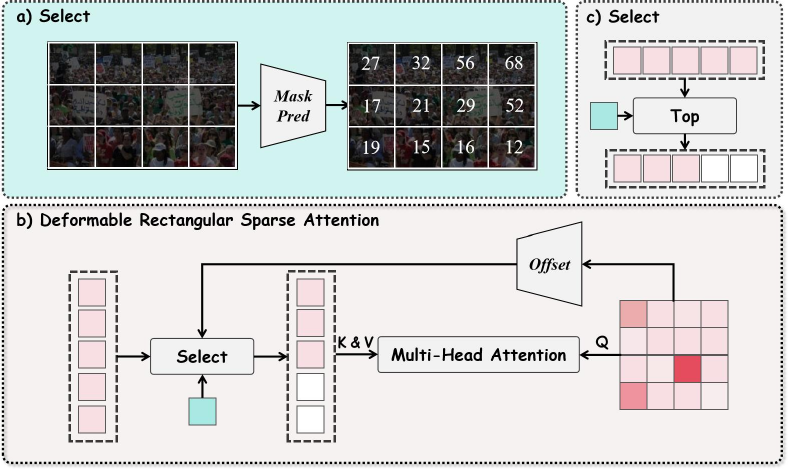}
    \caption{Overview of Deformable Rectangular Sparse Attention (DRSA).
(a) A mask prediction block estimates the maximum number anchors within each rectangular window, enabling adaptive sparse token selection.
(b) Based on the selected anchors, DRSA performs attention computation in a deformable rectangular manner: the feature map are used as queries, and the top selected anchors are corresponding keys and values for efficient attention aggregation.
(c) The number of retained anchors is dynamically determined for each window according to the predicted anchor quantity.}
    \label{fig:DRSA}
\end{figure}

Constructing a full hyper-graph over the entire feature map is computationally expensive for dense crowd scenes. Therefore, we perform hyper-graph reasoning within non-overlapping local windows. Specifically, $F^{fuse}$ is padded when necessary and divided into windows of size $S \times S$. For each local window, the feature is reshaped as:
\begin{equation}
X \in \mathbb{R}^{B \times N \times C}, \qquad N=S^2,
\end{equation}
where each token corresponds to one spatial location. Similarly, the foreground confidence is reshaped as:
\begin{equation}
W \in \mathbb{R}^{B \times N \times 1}.
\end{equation}

To construct the hyper-graph, we first normalize the window features and compute pairwise cosine similarity:
\begin{equation}
A = XX^{\top},
\end{equation}
Based on the confidence of the foreground $W$ and the similarity matrix of window features $A$, the nodes are divided into coarse foreground and background groups. Specifically, a node is treated as foreground if its confidence is larger than the mean confidence within the current window. This yields the class mask that reduces noisy cross group relations.

For each node, we select its five most similar neighbors within the same coarse group to construct hyperedges. Let $H \in \mathbb{R}^{N \times E}$ denote the incidence matrix of the hypergraph, where
\begin{equation}
H(u,e)=
\begin{cases}
1, & \text{if node } u \text{ is incident to hyperedge } e,\\
0, & \text{otherwise}.
\end{cases}
\end{equation}
In particular, each node acts as the center of a hyperedge and is connected with its top-5 most similar neighbors, thereby modeling local high-order relations.

After constructing the incidence matrix $H$, we perform normalized hyper-graph propagation. Let $D_v$ and $D_e$ denote the node-degree and hyperedge-degree matrices, respectively. The normalized hyper-graph operator is formulated as
\begin{equation}
\Theta = D_v^{-\frac{1}{2}} H D_e^{-1} H^{\top} D_v^{-\frac{1}{2}}.
\end{equation}
The propagated feature is then written as
\begin{equation}
X^{hyper} = \sigma(\Theta X W_h),
\end{equation}
where $W_h$ is a learnable linear projection. Finally, a residual connection and normalization layer are adopted:
\begin{equation}
X^{out} = \mathrm{Norm}\big(\mathrm{ReLU}(X^{hyper} + X)\big).
\end{equation}
Through such hyper-graph message passing, nodes can aggregate complementary information from multiple related neighbors simultaneously, rather than being restricted to pairwise fusion. This is particularly beneficial in low-light scenarios, where degraded RGB features can be progressively calibrated by stable geometry and boundary cues derived from depth and canny modalities.

After local hyper-graph reasoning, the window-wise features are restored to the original spatial layout and cropped to remove padded regions, yielding the final fused representation:
\begin{equation}
F^{mh}=F^{rgb} + \text{Reshape}(X^{out})
\end{equation}
The fused feature is then fed into the subsequent Deformable Rectangular Sparse Attention module for efficient dense crowd representation learning.

\subsection{Granularity Filtering}
Given the hypergraph-fused feature map $F^{mh} \in \mathbb{R}^{B \times C \times H \times W}$, we first apply average pooling and partition it into rectangular regions, resulting in a feature representation of size $\mathbb{R}^{B \times C \times \lceil H/h \rceil \times \lceil W/w \rceil}$, where $\lceil \cdot \rceil$ denotes the ceiling operation, and $h$ and $w$ denote the height and width of each rectangle, respectively. Then, a convolutional layer followed by a Sigmoid activation is employed to predict a granularity score for each rectangular region, denoted as $N^{gf} \in \mathbb{R}^{B \times 1 \times \lceil H/h \rceil \times \lceil W/w \rceil}$.
The maximum individual prediction function $\mathcal{M}(\cdot)$ is formulated as:
\begin{equation}
    \hat{n}_i = \mathcal{M}(F_i) = \varphi_{\max} * N^{gf}_i, \quad \hat{n}_i \in \mathbb{R}^{B \times 1},
\end{equation}
where $i \in [0, N_w - 1]$, and $N_w = \lceil H/h \rceil \cdot \lceil W/w \rceil$ denotes the total number of rectangular regions.

\subsection{Deformable Rectangular Sparse Decoder}
The Deformable Rectangular Sparse Decoder is an important component in LCNet which is consists of two Deformable Rectangular Sparse Attention (DRSA) modules.
As shown in Fig.~\ref{fig:DRSA}, the DRSA module improves the efficiency of dense crowd prediction by dynamically reducing the number of selected Keys and Values. Specifically, the DRSA module estimates the number of informative anchors within each rectangular window via granularity filtering and a mask prediction block, and subsequently applies sparse deformable attention only to the selected anchors.

Specifically, given the fused feature map $F \in \mathbb{R}^{B\times C \times H \times W}$, we divide it into a set of rectangular windows that do not overlap. For each window, a mask prediction block estimates the maximum number of target-related anchors to be preserved:
\begin{equation}
\hat{n}_i = \mathcal{M}(F_i),
\end{equation}
where $F_i \in \mathbb{R}^{B \times h\times w \times C}$ denotes the local characteristic in the rectangular window $i$-th, and $\hat{n}_i$ is the predicted anchor number for this window. In this way, different windows can adaptively retain different numbers of anchors according to their local crowd complexity.

Based on the predicted anchor number, we dynamically select the top anchors in each window. The dynamic number of retained tokens is obtained from the granularity score through
\begin{equation}
\hat{n}_i = \mathrm{clamp}\big(\lceil \sigma(s_i)\cdot \varphi_{\max}\rceil,\; 1,\; \varphi_{\max}\big),
\end{equation}
where $s_i$ is the predicted granularity logit of the $i$-th window, $\sigma(\cdot)$ is the Sigmoid function, $\lceil\cdot\rceil$ is the ceiling operation, and $\varphi_{\max}$ is the maximum number of anchors allowed per window. The token indices are then selected by
\begin{equation}
\mathcal{I}_i = \mathrm{Top}(P_i, \hat{n}_i),
\end{equation}
where $P_i$ denotes the tuple of anchor scores and positions and $\mathcal{I}_i$ denotes the set of selected anchors in the $i$-th window. This dynamic filtering strategy allows dense regions to preserve more informative anchors, while simple or low-confidence regions only keep a few representative ones, thereby reducing redundant computation.

\noindent\textbf{Differentiability.} The selection above involves $\lceil\cdot\rceil$, integer casting, and hard top-$\hat{n}_i$ masking, which are non-differentiable and block gradients to the anchor-number predictor. We therefore do not back-propagate through the discrete count: the granularity predictor is supervised directly by the granularity filtering loss $\mathcal{L}_{gf}$ (Sec.~\ref{sec:loss}) using the ground-truth per-window count as a soft target, while the selected tokens form a standard differentiable attention sub-graph through which gradients flow normally. This decoupling keeps training stable while still enabling adaptive sparsity at inference.

Then, the selected anchors are used for deformable rectangular sparse attention. Following deformable attention, we do not form keys from a full key projection; instead, given the query feature $Q_i$, we directly predict per-head sampling offsets and attention logits from the query:
\begin{equation}
\Delta P_i = \mathrm{MLP}_{off}(Q_i), \qquad
A_i = \mathrm{MLP}_{attn}(Q_i),
\end{equation}
and obtain the deformable sampling positions by
\begin{equation}
P^{sample}_i = P^{ref}_i + \Delta P_i,
\end{equation}
where $P^{ref}_i$ denotes the reference positions in the $i$-th rectangular window. The value features are sampled at these continuous positions via bilinear grid sampling:
\begin{equation}
V_i = \mathrm{GridSample}(F_i, P^{sample}_i).
\end{equation}
The output is the attention-weighted aggregation of the sampled values, where the attention weights are obtained by normalizing the predicted logits over the selected anchors:
\begin{equation}
\mathrm{Attn}(Q_i) = \sum_{k \in \mathcal{I}_i} \mathrm{Softmax}_k\!\big(A_i\big)\, V_{i,k}.
\end{equation}

Unlike standard scaled dot-product attention, the attention weights here are predicted directly from the query rather than computed from explicit query-key inner products, which avoids materializing the full $QK^{\top}$ similarity matrix. Together with the dynamic anchor selection, the DRSA reduces the number of query-anchor interactions and concentrates computation on informative foreground regions.


\begin{table*}[t]
\centering
\caption{Comparison with state-of-the-art methods on SHA\_Dark, SHB\_Dark, and LC-Crowd datasets. The best results are in \textbf{bold}, and the second-best results are underlined.}
\label{tab:sota_compare}
\resizebox{\tabsota}{!}{
\begin{tabular}{l|ccc|ccc|ccc}
\hline
\multirow{2}{*}{Method} & \multicolumn{3}{c|}{SHA\_Dark} & \multicolumn{3}{c|}{SHB\_Dark} & \multicolumn{3}{c}{LC-Crowd} \\
 & MAE$\downarrow$ & MSE$\downarrow$ & NAE$\downarrow$ & MAE$\downarrow$ & MSE$\downarrow$ & NAE$\downarrow$ & MAE$\downarrow$ & MSE$\downarrow$ & NAE$\downarrow$ \\
\hline
CSRNet \cite{CSRNet}     & 251.62 & 374.93 & 0.6767 & 93.47 & 132.61 & 0.6277 & 31.35 & 82.48 & 0.4285 \\
SANet  \cite{SASNet}     & 114.99 & 187.19 & 0.3577 & 35.31 & 59.58  & 0.2719 & 31.33 & 83.72 & 0.8466 \\
CAN  \cite{CANNet}       & 110.58 & 178.10 & 0.3028 & 29.31 & 49.14  & 0.2488 & 31.44 & 84.28 & 0.8412 \\
P2PNet  \cite{P2PNet}    & 57.85  & 96.52  & 0.1363 & 10.04 & 17.53  & 0.0700 & 28.28 & 79.95 & 0.4328 \\
PET \cite{PET}        & 57.28 & 95.73 & 0.1381 & \underline{8.25} & 14.53 & \underline{0.0600} & \underline{14.78} & 61.37 & \textbf{0.2371} \\
AWCC-Net \cite{awcc}   & 252.52 & 377.82 & 0.6600 & 69.03 & 104.15 & 0.6457 & 31.77 & 84.98 & 0.9283 \\
STEERER   \cite{STEERER}  & 84.66  & 130.55 & 0.2100 & 13.57 & 24.37  & 0.1034 & 36.70 & 85.73 & 0.8531 \\
APGCC   \cite{APGCC}    & 59.20  & 92.50  & 0.1477 & 8.81  & \underline{13.81}  & 0.0782 & 15.16 & 59.23 & \underline{0.2714} \\
FGENet  \cite{FGENet}    & \underline{53.58}  & 95.25  & \underline{0.1234} & 8.95  & 14.42  & 0.0722 & 15.56  & \textbf{27.75} & 0.4061 \\
M2PLNet  \cite{M2PLNet}    & 55.53  & \underline{90.85}  & 0.1282 & 18.57  & 28.43  & 0.1496 & 24.43  & 44.31 & 0.6378 \\
\hline
\textbf{LCNet (Ours)} & \textbf{51.08} & \textbf{82.87} & \textbf{0.1205} & \textbf{6.55} & \textbf{10.36} & \textbf{0.0550} & \textbf{12.20} & \underline{36.93} & 0.3747 \\
\hline
\end{tabular}
}
\end{table*}
\subsection{Loss Function}\label{sec:loss}

The overall loss consists of three parts: the \emph{classification loss}, the \emph{localization loss}, and the \emph{granularity filtering loss}. The final objective is formulated as:
\begin{equation}
\mathcal{L} = \lambda_c \mathcal{L}_{cls} + \lambda_r \mathcal{L}_{reg} + \lambda_w \mathcal{L}_{gf},
\end{equation}
where $\lambda_c$, $\lambda_r$, and $\lambda_w$ are the balancing coefficients for the three loss terms.

\paragraph{\textbf{Classification and localization losses}}
Given the predicted anchors and the ground-truth crowd points, we first perform bipartite matching using the Hungarian algorithm \cite{kuhn1955hungarian} to assign each prediction to a ground-truth target. Based on the matched pairs, the classification loss is computed by Cross-Entropy, which supervises whether each anchor belongs to the foreground or background. The localization loss is computed by Smooth L1 loss \cite{smoothl1} on the predicted point coordinates. Therefore, the point prediction losses are defined as
\begin{equation}
\mathcal{L}_{cls} = \mathrm{CE}(\hat{y}, y),
\end{equation}
\begin{equation}
\mathcal{L}_{reg} = \mathrm{Smooth_{L1}}(\hat{p}, p),
\end{equation}
where $\hat{y}$ and $y$ denote the predicted and target classification labels and $\hat{p}$ and $p$ denote the predicted and ground-truth point coordinates, respectively.

\paragraph{\textbf{Granularity filtering loss}}
To supervise the proposed \emph{Granularity Filtering} module, we introduce an auxiliary loss on the predicted granularity map $N^{gf}$. Specifically, for each rectangular window, we compute the absolute error between the predicted number of anchors and the target number derived from the ground-truth crowd annotations within that window. Let $\varphi_{\max}$ denotes the maximum anchor number allowed in each window, and let $N_w$ denote the total number of windows. The granularity filtering loss is defined as:
\begin{equation}
\mathcal{L}_{gf} = \frac{1}{N_w \cdot \varphi_{\max}} \sum_{i=1}^{N_w} \left| \hat{n}_i - n_i \right|,
\end{equation}
where $\hat{n}_i = \varphi_{\max} \cdot N^{gf}_i$ is the predicted number of anchors for the $i$-th window, and $n_i$ is the corresponding number of ground-truth anchor.

\section{Experiments}

\subsection{Experimental Settings}
\textbf{Training Details.}
All experiments are conducted under the PyTorch 2.0.1 framework with CUDA 11.6, using an NVIDIA V100 GPU with 32GB memory. The pre-trained VGG16-bn \cite{vgg} is adopted as the backbone network. The model is optimized using the Adam \cite{adam} optimizer with a learning rate of $1 \times 10^{-5}$, the random seed is fixed to 42.

\textbf{Hyperparameter Settings:} 
For the construction of hypergraph nodes, the parameter $S$ is set to 8. The height $h$ and width $w$ of the rectangular window are set to 8 and 16, respectively. The maximum number $\varphi_{max}$ of anchor points is set to 128. The weight $\lambda_r$ is set to 0.05, the hyperparameter for window prediction anchors  $\lambda_w$ is 5, and the confidence loss weight  $\lambda_c$ is set to 1.0.

\textbf{Data Augmentation:} 
We randomly crop $256 \times 256$ patches from each image as training samples, and apply random scaling and random flipping. For datasets containing high-resolution images, we resize the images while maintaining their original aspect ratios. Specifically, the maximum length of the longer side is restricted to not exceed 1408 and 1920 (for the respective dataset), strictly keeping the original aspect ratio.


\subsection{Main Results}
As shown in Table~\ref{tab:sota_compare}, LCNet achieves the best overall performance across the three low-light crowd counting benchmarks. In particular, on both SHA\_Dark and SHB\_Dark, our method consistently outperforms all competing approaches in terms of MAE, MSE, and NAE, demonstrating its strong robustness under severe synthetic illumination degradation. Compared with the second-best method, LCNet reduces the MAE/MSE by 4.7\%/8.8\% on SHA\_Dark and 20.6\%/25.0\% on SHB\_Dark, respectively. These results indicate that LCNet can effectively preserve discriminative crowd representations and maintain reliable counting performance under extremely low-light conditions.
On the real-world LC-Crowd dataset, LCNet also achieves the best MAE, surpassing PET by 17.5\% and 39.8\%, respectively. The particularly large gain in MSE suggests that our method yields more stable predictions in complex real-world low-light scenes. Although LCNet does not obtain the best NAE on LC-Crowd, it still achieves the most competitive overall performance, indicating that the proposed framework generalizes effectively from synthetic low-light benchmarks to real-world dark environments.
In addition, we further evaluate our framework against strong multi-modal baselines on the dark part of RGBT-CC in Sec.~\ref{exp:rgbtcc}, and compare with the conventional enhance-then-count strategy in Table~\ref{tab:comparison_sha_dark}. These experiments further demonstrate that our method still achieves the best performance under multi-modal low-light conditions.

\noindent\textbf{On comparison fairness.} Since LCNet takes RGB together with depth and Canny priors while most baselines in Table~\ref{tab:sota_compare} are RGB-only, the main table alone cannot fully isolate the architectural contribution from the effect of the extra inputs. We address this from two complementary angles. First, the fusion-strategy ablation in Table~\ref{tab:fusion_strategy} feeds \emph{all} variants the identical RGB+Depth+Canny input and shows that our hyper-graph reasoning still outperforms simpler fusion operators, indicating the gain is not merely from extra modalities. Second, the multi-modal comparison on RGBT-CC-Dark (Sec.~\ref{exp:rgbtcc}) evaluates against strong multi-modal baselines under the same input protocol.

\begin{table}[htbp]
\centering
\caption{Comparison of model size and runtime. ``\#Params'' is in millions and GFLOPs is measured at a fixed input resolution of $1024\times768$.}
\label{tab:model_comparison}
\begin{tabular}{lccc}
\toprule
Method  & Params (M) & GFLOPs & FPS \\
\midrule
PET      & 20.909  & --  & 9.31  \\
FGENet    & 183.271 & -- & 19.46 \\
M2PLNet  & 508.461 & -- & 7.68  \\
Ours     & 22.590  & -- & 5.96  \\
\bottomrule
\end{tabular}
\end{table}
As shown in Table~\ref{tab:model_comparison}, our method contains 22.590M parameters, which is very close to PET (20.909M) and substantially smaller than FGENet (183.271M) and M2PLNet (508.461M). This indicates that LCNet remains lightweight in parameter size. We note that its measured inference speed (5.96 FPS) is currently lower than the compared methods. We attribute this gap to implementation overhead rather than to the algorithmic complexity of DRSA: the dynamic top-$k$ selection and the bilinear grid sampling involve data-dependent indexing and scatter/gather operations that are not yet fused into GPU-friendly kernels, which dominate the wall-clock time despite the reduced number of attention interactions. Therefore, we position DRSA as a \emph{parameter-efficient and computation-adaptive} design that concentrates attention on informative foreground regions, rather than as a runtime-fastest module, and we leave kernel-level optimization of the sparse sampling operators as future work. We have correspondingly toned down the efficiency claims throughout the paper.

\subsection{Ablation Studies}\label{sec:ablation}
\begin{table}[htbp]
\centering
\begin{minipage}[t]{\tabhalf}
\centering
\caption{Comparison of different attention modules on SHA\_Dark.}
\label{tab:attention_module}
\resizebox{\linewidth}{!}{
\begin{tabular}{lccc}
\hline
Module & MAE$\downarrow$ & MSE$\downarrow$ & NAE$\downarrow$ \\
\hline
Rectangular Attention \cite{PET}        & 57.28 & 95.73 & 0.1381 \\
Deformable Attention    \cite{defatten}      & 53.36 & 89.36 & 0.1238 \\
Sparse Rectangular Attention & 53.36 & 89.36 & 0.1238 \\
Dilateformer \cite{dilateformer}            & 53.49 & 87.49 & 0.1261 \\
\hline
DRSA         & \textbf{51.08} & \textbf{82.87} & \textbf{0.1205} \\
\hline
\end{tabular}
}
\end{minipage}
\hfill
\begin{minipage}[t]{\tabhalf}
\centering
\caption{Ablation study of different fusion positions on SHA\_Dark (see text for the meaning of each position).}
\label{tab:fusion_position}
\resizebox{\linewidth}{!}{
\begin{tabular}{lccc}
\hline
Fusion Position & MAE$\downarrow$ & MSE$\downarrow$ & NAE$\downarrow$ \\
\hline
Attention map & 53.20 & 85.82 & 0.1253 \\
Value branch & 53.03 & 85.00 & 0.1249 \\
FPN only & 52.94 & 84.32 & 0.1312 \\
Both & 54.46 & 87.52 & 0.1310 \\
\hline
\textbf{Ours} &  \textbf{51.08} & \textbf{82.87} & \textbf{0.1205} \\
\hline
\end{tabular}
}
\end{minipage}
\end{table}
\paragraph{\textbf{Ablation Studies on different attention modules}}
As shown in Table~\ref{tab:attention_module}, the proposed DRSA achieves the best MAE and MSE among all compared attention modules, demonstrating its superiority in improving overall counting accuracy and prediction stability on the SHA\_Dark dataset. Compared with the basic Rectangular Attention, DRSA reduces the MAE from 57.28 to 51.08 and the MSE from 95.73 to 82.87, indicating that the proposed design can better capture discriminative crowd representations under low-light conditions. Although DRSA does not obtain the best NAE, its consistent gains in MAE and MSE suggest that it is more effective for optimizing overall counting performance, while the relative error on sparse samples still has room for further improvement.

\paragraph{\textbf{Ablation Studies on different fusion positions}}
As shown in Table~\ref{tab:fusion_position}, the fusion position has a clear impact on the final performance, indicating that the effectiveness of auxiliary modalities depends not only on the introduced information itself but also on where it is fused. Here, ``attention map'' injects the auxiliary feature as a mask into the Transformer attention computation; ``value branch'' performs fusion on the value branch in a Retinexformer-like manner \cite{retinexformer}; ``FPN only'' applies fusion only in the FPN; and ``both'' applies fusion in the FPN and the Deformable Rectangular Sparse Decoder. Among the Transformer-internal strategies, the value-branch design consistently performs better than the attention-map design, suggesting that directly enriching feature representations is more effective than merely modulating attention weights. The FPN-only strategy achieves the best MSE, indicating that late-stage fusion is more beneficial for suppressing large errors on hard samples. In contrast, applying fusion in both the FPN and the Deformable Rectangular Sparse Decoder leads to degraded performance, implying that auxiliary cues may introduce redundancy and feature interference. Our method achieves the best MAE, demonstrating its superiority in improving the overall counting accuracy.

\begin{table}[htbp]
\centering
\begin{minipage}[t]{\tabhalf}
\centering
\caption{Ablation of different modality combinations on SHA\_Dark.}
\label{tab:modality_ablation}
\resizebox{\linewidth}{!}{
\begin{tabular}{lccc}
\hline
Modal & MAE$\downarrow$ & MSE$\downarrow$ & NAE$\downarrow$ \\
\hline
RGB & 57.28 & 95.73 & 0.1381 \\
RGB+Depth & 52.89 & 86.33 & 0.1253 \\
RGB+Canny & 51.77 & \textbf{81.54} & 0.1287 \\
\hline
RGB+Depth+Canny & \textbf{51.08} & 82.87 & \textbf{0.1205} \\
\hline
\end{tabular}
}
\end{minipage}
\hfill
\begin{minipage}[t]{\tabhalf}
\centering
\caption{Ablation of the fusion strategy under the same RGB+Depth+Canny input on SHA\_Dark.}
\label{tab:fusion_strategy}
\resizebox{\linewidth}{!}{
\begin{tabular}{lccc}
\hline
Fusion Strategy & MAE$\downarrow$ & MSE$\downarrow$ & NAE$\downarrow$ \\
\hline
Simple Concat            & \texttt{TBD} & \texttt{TBD} & \texttt{TBD} \\
Gated Fusion             & \texttt{TBD} & \texttt{TBD} & \texttt{TBD} \\
Cross-Attention Fusion   & \texttt{TBD} & \texttt{TBD} & \texttt{TBD} \\
Pairwise Graph (GCN)     & \texttt{TBD} & \texttt{TBD} & \texttt{TBD} \\
w/o FG/BG grouping & \texttt{TBD} & \texttt{TBD} & \texttt{TBD} \\
\hline
\textbf{Hyper-Graph (Ours)} & \textbf{51.08} & \textbf{82.87} & \textbf{0.1205} \\
\hline
\end{tabular}
}
\end{minipage}
\end{table}
As shown in Table~\ref{tab:modality_ablation}, the single-modality RGB baseline exhibits the highest counting errors, indicating that relying solely on RGB information is insufficient for accurate crowd counting under low-light conditions. Incorporating depth information reduces both MAE and MSE significantly, suggesting that depth provides valuable geometric cues for improving count estimation in dense regions. Introducing edge (canny) information further lowers the MSE, demonstrating that edge features help stabilize predictions on challenging samples. Finally, fusing RGB, depth, and edge modalities achieves the best performance in terms of MAE and NAE, confirming that the complementary information from multiple modalities substantially enhances counting accuracy and reduces relative errors in low-light scenarios.

\paragraph{\textbf{Ablation Studies on the fusion strategy}}
To isolate the contribution of the proposed high-order reasoning, we compare our Hyper-Graph Fusion against several pairwise or first-order alternatives under the \emph{same} RGB+Depth+Canny input, as reported in Table~\ref{tab:fusion_strategy}. Simple concatenation and gated fusion combine the cues without explicit relational reasoning; cross-attention and a pairwise graph (GCN) model only second-order (node-to-node) relations; and the ``w/o FG/BG grouping'' variant removes the foreground/background constraint when building hyperedges. The results show that our hyper-graph design consistently outperforms these alternatives, confirming that the gain originates from the high-order, group-wise calibration rather than merely from the extra depth and edge inputs, and that the foreground/background grouping is important for suppressing noisy cross-group connections.

\begin{figure}[t]
    \centering
    \begin{subfigure}[b]{0.49\linewidth}
        \centering
        \includegraphics[width=\linewidth]{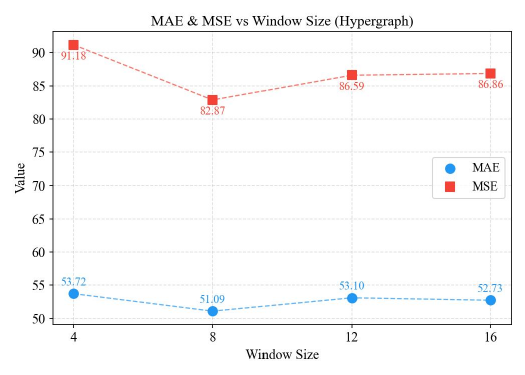}
        \caption{Comparison of MAE and MSE on SHA\_Dark under different hypergraph window sizes.}
        \label{fig:window_size}
    \end{subfigure}
    \hfill
    \begin{subfigure}[b]{0.49\linewidth}
        \centering
        \includegraphics[width=\linewidth]{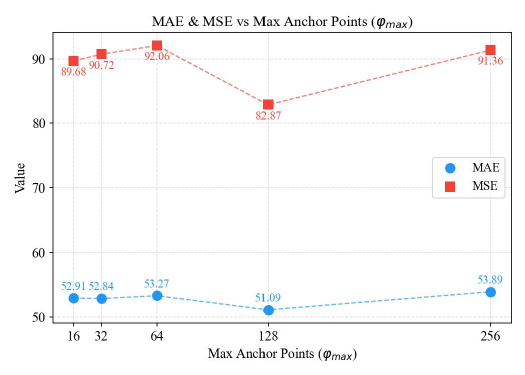}
        \caption{Comparison of MAE and MSE on SHA\_Dark under different max anchor points $\varphi_{max}$.}
        \label{fig:anchor_points}
    \end{subfigure}
    \caption{Ablation study on SHA\_Dark with different hypergraph settings.}
    \label{fig:ablation_hypergraph}
\end{figure}

As shown in Fig.~\ref{fig:ablation_hypergraph}, both the hypergraph window size and the maximum number of anchor points have noticeable effects on performance. The best result for window size is achieved at 8, while increasing the maximum anchor points $\varphi_{max}$ generally improves the performance, with the best metrics obtained at 128.
\begin{figure}
    \centering
    \includegraphics[width=\figcol]{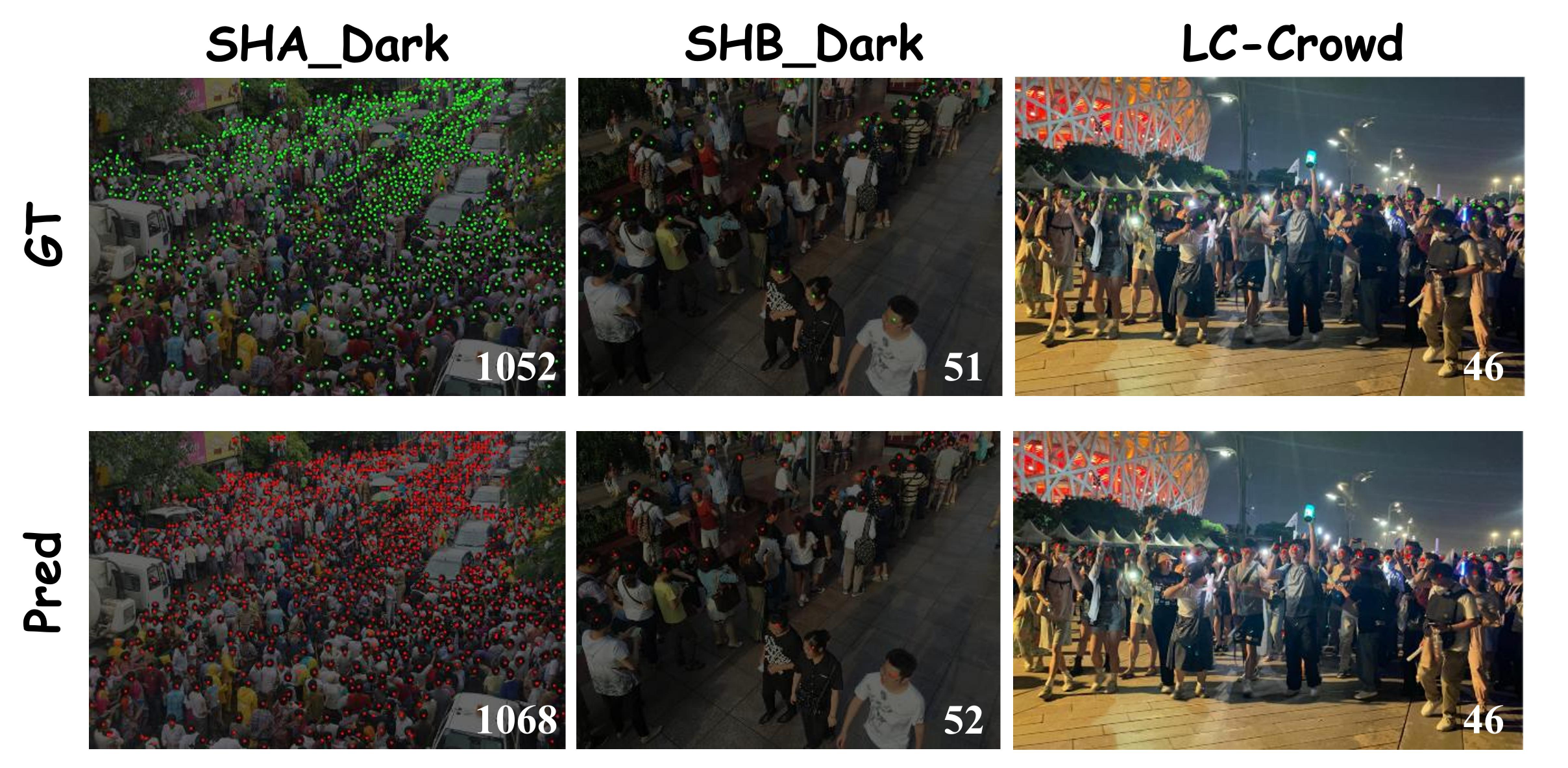}
    \caption{Visualization of crowd counting results on three benchmark datasets with different densities.}
    \label{fig:res_vis}
\end{figure}
\subsection{Qualitative Analysis}
\paragraph{\textbf{Visualization of prediction results}}
Fig.~\ref{fig:res_vis} presents qualitative results on three benchmark datasets. As shown in the figure, our method can accurately localize head positions and produce reliable counting estimates across diverse crowd scenes, including highly congested street scenes, queueing scenarios, and nighttime gathering scenes. Even in extremely dense regions, the predicted points remain well aligned with the underlying head distributions, demonstrating the robustness of our method in challenging cases. In relatively sparse scenes, the model still maintains precise localization ability without obvious missed detections or repeated predictions. These results further verify that our method generalizes well across datasets with varying crowd densities and scene characteristics.

\begin{figure}[h]
    \centering
    \includegraphics[width=\figcol]{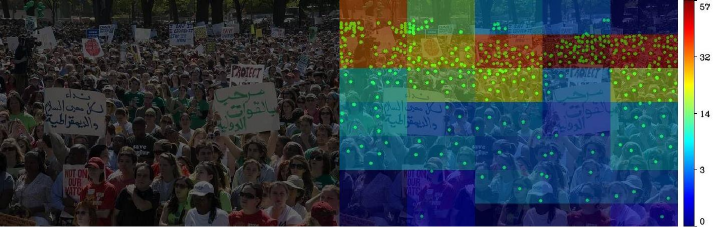}\vspace{-0.1cm}
    \caption{Visualization of anchor distribution in the proposed Deformable Rectangular Sparse Attention.}
    \label{fig:anchor_vis}
\end{figure}
\vspace{-0.1cm}
\paragraph{\noindent\textbf{Visualization of the anchor distribution in DRSA module.}}
 Fig.\ref{fig:anchor_vis} shows that the proposed DRSA module automatically allocates more anchors to crowd-dense regions while assigning fewer anchors to background areas. The left panel shows the input image, while the right panel presents the learned anchor responses overlaid on the image. Colored rectangular regions indicate the adaptively selected sparse attention windows, and the green points denote anchor locations. Warmer colors correspond to higher anchor responses, showing that the proposed attention mechanism focuses more on informative regions with dense crowd distributions. Meanwhile, the attention windows are adaptively adjusted in a rectangular manner to better fit the spatial layout of the crowd. This demonstrates that the proposed module can focus computation on informative regions and effectively reduce redundant computation.

\subsection{Comparison with Enhance-then-Count Pipelines}
A natural alternative to our task-specific design is to first enhance the low-light image and then apply a standard counter. To verify that explicit reflectance re-calibration is more effective than such a decoupled strategy, we compare three settings on SHA\_Dark in Table~\ref{tab:comparison_sha_dark}: the plain \textbf{Baseline} counter, a \textbf{Seq. Pipeline} that enhances the input with Zero-DCE \cite{Zero-DCE} before counting, and \textbf{Ours}.

\begin{table}[htbp]
\centering
\caption{Comparison of different pipelines on SHA\_Dark. ``Seq. Pipeline'' first enhances the image with Zero-DCE and then feeds it to the baseline counter.}
\label{tab:comparison_sha_dark}
\begin{tabular}{lccc}
\hline
Method & MAE$\downarrow$ & MSE$\downarrow$ & NAE$\downarrow$ \\
\hline
Baseline      & 57.28 & 95.73 & 0.1381 \\
Seq. Pipeline & 53.40 & 84.97 & 0.1300 \\
Ours          & \textbf{51.08} & \textbf{82.87} & \textbf{0.1205} \\
\hline
\end{tabular}
\end{table}
The \textbf{Seq. Pipeline} improves over the \textbf{Baseline} on all three metrics (MAE $57.28\!\rightarrow\!53.40$, MSE $95.73\!\rightarrow\!84.97$, NAE $0.1381\!\rightarrow\!0.1300$), confirming that mitigating low-light degradation is beneficial. However, it remains clearly inferior to \textbf{Ours} ($51.08$/$82.87$/$0.1205$). A likely reason is that off-the-shelf enhancers such as Zero-DCE are optimized for visual quality rather than for preserving the semantic and structural cues that dense counting relies on, so the cascaded design may distort features or fail to recover informative details in heavily underexposed regions. In contrast, our method re-calibrates degraded reflectance features directly with illumination-robust priors, which is better aligned with the counting objective.

\subsection{Comparison with Multi-Modal Methods on RGBT-CC-Dark}
\label{exp:rgbtcc}
To further assess our framework against strong \emph{multi-modal} baselines under a unified input protocol, we evaluate on the dark split of RGBT-CC \cite{IADM}. Since this benchmark provides thermal rather than depth as the auxiliary modality, we instantiate two variants of our model: \textbf{Ours (T)} uses the paired thermal image as the auxiliary cue, while \textbf{Ours (Depth)} uses an estimated depth map. The results are reported in Table~\ref{tab:rgbtcc_results}.

\begin{table}[htbp]
\centering
\caption{Comparison with multi-modal methods on the RGBT-CC-Dark dataset. Best in \textbf{bold}, second-best \underline{underlined}.}
\label{tab:rgbtcc_results}
\begin{tabular}{lcc}
\toprule
 Method & MAE$\downarrow$ & RMSE$\downarrow$ \\
\midrule
 IADM \cite{IADM}           & 15.44 & 29.11 \\
 TAFNet \cite{TAFNet}        & 14.20 & 27.50 \\
 GETANet \cite{GETANet}       & 12.93 & 22.63 \\
 MIANet \cite{MIANet}         & 13.88 & 25.15 \\
 DEFNet \cite{DEFNet}         & 12.39 & 23.16 \\
 CMFX \cite{CMFX}         & 12.17 & 21.66 \\
 CMPNet \cite{CMPNet}        & 10.46 & 17.79 \\
 RGBT-Booster \cite{rgbtBooster}  & 10.35 & 18.15 \\
 MISF-Net \cite{misfnet}     & 10.13 & 17.96 \\
 Niu et al. \cite{niuetc}            & \underline{9.81} & \underline{17.06} \\ \hline
 Ours (T)               & \textbf{9.74} & \textbf{14.56} \\
 Ours (Depth)               & 11.47 & 19.11 \\
\bottomrule
\end{tabular}
\end{table}
With the thermal cue, \textbf{Ours (T)} achieves the best performance (MAE $9.74$, RMSE $14.56$), surpassing the previous best method by Niu et al. ($9.81$/$17.06$); the particularly large RMSE reduction indicates stronger suppression of large estimation errors. This confirms that our hyper-graph re-calibration generalizes to other illumination-robust modalities under the same input protocol, rather than being tied to depth. We also observe that \textbf{Ours (Depth)} ($11.47$/$19.11$) trails \textbf{Ours (T)} and several competitors, suggesting that in fully dark RGBT-CC scenes thermal provides more discriminative target responses than estimated depth. This is consistent with our motivation that no single auxiliary cue is universally reliable, which is precisely why high-order, group-wise calibration over multiple cues is desirable.

\section{Conclusion}
This paper addresses the practical but insufficiently studied problem of crowd counting in low-light environments. Departing from the conventional enhance-then-count pipeline, we recast the task from a Retinex perspective as \emph{reflectance re-calibration}, recovering illumination-robust crowd representations at the feature level rather than restoring a visually pleasing image at the pixel level. Guided by this view, we introduce depth and edge cues as illumination-invariant calibration anchors, and observe that any single cue may fail locally under extreme darkness. We therefore propose a Multi-Modal Hyper-Graph Fusion module that performs high-order, group-wise calibration over appearance, geometry, and structure cues, going beyond fragile pairwise fusion. Since the calibrated foreground remains spatially sparse in nighttime scenes, we further design a Deformable Rectangular Sparse Attention module that adaptively allocates computation to informative regions. To support systematic study, we build three new benchmarks, namely SHA\_Dark, SHB\_Dark, and LC-Crowd, covering both synthetic and real-world scenarios. Extensive experiments demonstrate that the proposed LCNet achieves the best overall performance, outperforming previous state-of-the-art approaches on most metrics across all three datasets. We also discuss the current runtime limitation of the sparse sampling operators and identify kernel-level optimization of these operators as a clear direction for future work.

\bibliographystyle{ACM-Reference-Format}
\bibliography{acmart}
\appendix
\section{Evaluation Metrics}
To comprehensively evaluate the counting performance, we adopt three widely used metrics, namely Mean Absolute Error (MAE), Mean Squared Error (MSE), and Normalized Absolute Error (NAE). These metrics measure the counting accuracy from different perspectives, including absolute deviation, robustness to large errors, and relative estimation bias.

Specifically, MAE measures the average absolute difference between the predicted count and the ground-truth count:
\begin{equation}
\mathrm{MAE} = \frac{1}{N}\sum_{i=1}^{N} \left| \mathrm{Pred}_i - \mathrm{GT}_i \right|,
\end{equation}
where $N$ denotes the number of test images, $\mathrm{GT}_i$ denotes the ground-truth count of the $i$-th image, and $\mathrm{Pred}_i$ denotes the corresponding predicted count.

MSE is defined as:
\begin{equation}
\mathrm{MSE} = \sqrt{ \frac{1}{N}\sum_{i=1}^{N} \left( \mathrm{Pred}_i - \mathrm{GT}_i \right)^2 },
\end{equation}
which is more sensitive to large prediction errors and thus reflects the robustness of the model.

In addition, NAE evaluates the relative counting error with respect to the ground-truth count:
\begin{equation}
\mathrm{NAE} = \frac{1}{N}\sum_{i=1}^{N} \frac{\left| \mathrm{Pred}_i - \mathrm{GT}_i \right|}{\mathrm{GT}_i}.
\end{equation}

In all cases, lower values of MAE, MSE, and NAE indicate better counting performance.

\section{Additional Implementation Details}
\subsection{Modality Embedding}
For the depth and canny modalities, we employ two separate parallel multi-scale convolutional neural networks, each implemented by stacking three Conv-BN-ReLU layers to extract modality-specific features at different receptive fields.

\subsection{Prediction Head}
We use a single-layer MLP head to predict the classification result, indicating whether each existing anchor point belongs to the foreground or the background, formulated as $\sigma(\mathrm{Linear}(F_{\text{final}}))$. Anchor locations have already been propagated forward during the deformable rectangular sparse decoder stage.

\section{RGBT-CC-Dark Dataset Statistics}
For completeness, Table~\ref{tab:rgbtcc_statistics} summarizes the statistics of the RGBT-CC-Dark dataset used in the multi-modal comparison of Sec.~\ref{exp:rgbtcc}. It contains 1,017 images with 68,990 total annotations, split into 520/103/394 images for training/validation/testing. The average crowd count per image is 67.8 with a median of 49.0, ranging from 8 to 535, indicating considerable variation in crowd density across splits.

\begin{table}[htbp]
\centering
\caption{RGBT-CC-Dark dataset statistics.}
\label{tab:rgbtcc_statistics}
\begin{tabular}{lcccc}
\hline
 & Train & Val & Test & Total \\
\hline
Images            & 520   & 103   & 394   & 1,017 \\
Total Annotations & 32,510 & 6,977 & 29,503 & 68,990 \\
Average           & 62.5  & 67.7  & 74.9  & 67.8 \\
Median            & 51.0  & 52.0  & 44.0  & 49.0 \\
Min               & 10    & 12    & 8     & 8 \\
Max               & 535   & 523   & 406   & 535 \\
Std               & 48.9  & 74.1  & 83.2  & 67.0 \\
\hline
\end{tabular}
\end{table}

\end{document}